\documentclass[runningheads]{llncs}
\usepackage{mathptmx}
\usepackage{amssymb}
\usepackage{amsmath}
\usepackage{url}
\usepackage{graphicx}
\usepackage{multirow}
\usepackage{wrapfig}
\usepackage{hyperref}
\usepackage{color}
\usepackage{graphicx}
\usepackage{enumitem}
\usepackage{multicol}
\usepackage{bbm}
\usepackage{calrsfs}
\usepackage{booktabs}
\usepackage{stmaryrd}
\usepackage{CJK}
\usepackage[table]{xcolor}
\DeclareMathAlphabet{\pazocal}{OMS}{zplm}{m}{n}

\usepackage{hyperref}
\hypersetup{
	colorlinks=true,
	linkcolor=red,
	filecolor=red,      
	urlcolor=blue,
	citecolor=green,
}

\definecolor{Neoplastic}{RGB}{255, 0, 0}
\definecolor{Inflammatory}{RGB}{34, 221, 77}
\definecolor{Connective}{RGB}{35, 92, 235}
\definecolor{Dead}{RGB}{238, 233, 81}
\definecolor{Non-Neoplastic Epithelial}{RGB}{241, 162, 87}

\begin{document}
	\institute{Anonymous Submission}
	\title{Multi-Compound Transformer for Accurate  Biomedical Image Segmentation}
	\author{Yuanfeng Ji\inst{1}\and
				Ruimao Zhang\inst{2}\and
				Huijie Wang \inst{2}\and
				Zhen Li\inst{2} \and \\
				Lingyun Wu\inst{3} \and 
				Shaoting Zhang\inst{3} \and
				Ping Luo\inst{1}
				\thanks{Ping Luo is the corresponding author of this paper.}}
	\institute{The University of Hong Kong \and
		Shenzhen Research Institute of Big Data, The Chinese University of Hong Kong (Shenzhen) \and
		SenseTime Research
		\\}
	\maketitle
	\begin{abstract}
		The recent vision transformer (\textit{i.e.} for image classification) learns non-local attentive interaction of different patch tokens.
		However, prior arts miss learning the cross-scale dependencies of different pixels, the semantic correspondence of different labels, and the consistency of the feature representations and semantic embeddings, which  are critical for  biomedical segmentation.
		%
		%
		In this paper, we tackle the above issues by proposing a unified transformer network, termed Multi-Compound Transformer (MCTrans), which  incorporates rich feature learning and semantic structure mining into a unified framework.  
		Specifically, MCTrans embeds the multi-scale convolutional features as a sequence of tokens, and performs intra- and inter-scale self-attention, rather than single-scale attention in previous works.
		In addition, a learnable proxy embedding is also introduced to model semantic relationship and feature enhancement by using self-attention and cross-attention, respectively.  
		MCTrans can be easily plugged into a UNet-like network, 
		and attains a significant improvement over the state-of-the-art methods in biomedical image segmentation in six standard benchmarks.
		For example, MCTrans outperforms UNet by 3.64\%, 3.71\%, 4.34\%, 2.8\%, 1.88\%, 1.57\% in Pannuke, CVC-Clinic, CVC-Colon, Etis, Kavirs, ISIC2018 dataset, respectively.
		Code is available at \url{https://github.com/JiYuanFeng/MCTrans}.
		
	\end{abstract}
	\section{Introduction}
\begin{figure}[t!]
		\centering
		\includegraphics[width=\linewidth]{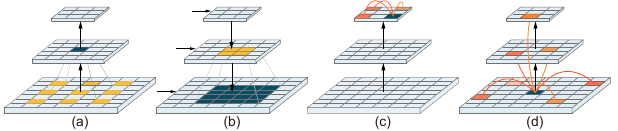}
		\caption{Conceptual comparison of various mechanisms for context modeling for segmentation. In contrast to (a-c), MCTrans models pixel-wise relationships between multiple scales features, enabling more consistent and effective context encoding. The Prussian blue grids denote the target pixel while other color grids represent the support pixels. For simplicity, we only show a subset of the pathways between target pixels and support pixels.}
		\label{fig:diff_arch}
	\end{figure}

	Medical image segmentation, which aims to automatically delineate anatomical structures and other regions of interest from medical images, is essential for modern computer-assisted diagnosis (CAD) applications, such as lesion detection \cite{codella2019skin,bernal2015wm,bernal2012towards,silva2014toward,jha2020sessile} and anatomical structure localization \cite{gamper2019pannuke}.
	Recent advances in segmentation accuracy are primarily driven by the power of convolution neural networks (CNN)~\cite{simonyan2014very,he2016deep}.
	%
	%
	However, due to the local property of the convolutional kernels, the traditional CNN-based segmentation models (\textit{e.g.} FCN~\cite{long2015fully}) lack the ability for modeling long-term dependencies.
	To address such an issue, various approaches have been exploited for powerful relation modeling.
	For example, the spatial pyramid based methods \cite{chen2017deeplab,zhao2017pyramid,gu2019net} adopt various sizes of convolutional kernels to aggregate contextual information from different ranges in a single layer (Fig.~\ref{fig:diff_arch} (a)).
	The UNet \cite{ronneberger2015u} based encoder-decoder networks \cite{ronneberger2015u,zhou2018unet++,ji2020uxnet}
	merge the coarse-grained deep features and fine-grained shallow features with the same scales by applying skip-connection.
	Although these methods achieved great success in dense prediction, it is still limited by the inefficient non-local context modeling among arbitrary positions,
	%
	%
	making it bleak for further promoting the accuracy of complex views.


	%
	%
	%
	%
	
	%
	Recently, the Vision Transformer~\cite{vaswani2017attention}, which is built upon learning attentive interaction of different patch tokens, has achieved much attention in various vision tasks \cite{dosovitskiy2020image,zhu2020deformable,carion2020end,xie2021segmenting}.
	For medical image segmentation, Chen \textit{et al.} firstly propose TransUNet~\cite{chen2021transunet}, which adopts the self-attention mechanism to compute global context at the highest-level CNN features, ensuring various ranges dependencies in a specific scale (Fig. \ref{fig:diff_arch} (c)).
	However, such a design is still sub-optimal for medical image segmentation for the following reasons.
	First, it only uses the self-attention mechanism for context modeling on a single scale but ignores the cross-scale dependency and consistency.
	The latter usually plays a critical role in the segmentation of lesions with dramatic size changes. 
	Second, beyond the context modeling, how to learn the correlation between different semantic categories and how to ensure the feature consistency of the same category region are still not taken into account.
	But both of them have become critical for CNN-based segmentation scheme design~\cite{yu2020context}.

	%
	%
	%
	%
	%
	%
	%
	%
	%
	
\begin{figure}[t!]
		\centering
		\includegraphics[width=\linewidth]{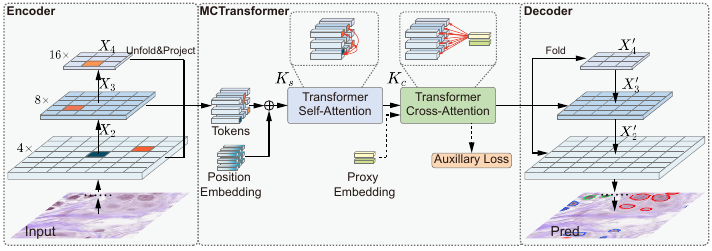}
		\caption{The overview of MCTrans. We use CNN to extract multi-scale features, and feed the embedded tokens to the Transformer-Self-Attention module to construct the multi-scale context. We add a learnable proxy embedding to learn category dependencies and interact with the feature representations via the Transformer-Cross-Attention module. Finally, we fold the encoded tokens to several 2D feature maps and merge them progressively to generate segmentation results. For the details of the two modules, please refer to Fig. \ref{fig:attention}.}
		\label{fig:overview}
	\end{figure}

	In this paper, attempting to overcome the limitations mentioned above, we propose the Multi-Compound Transformer (MCTrans), which incorporates rich context modeling and semantic relationship mining for accurate biomedical image segmentation.
	As illustrated in Fig. \ref{fig:overview}, MCTrans overcomes the limitations of conventional vision transformers by:
	(1) introducing the \textit{Transformer-Self-Attention} (TSA) module to achieve cross-scale pixel-level contextual modeling via the self-attention mechanisms, leading to a more comprehensive feature enhancement for different scales.
	(2) developing the \textit{Transformer-Cross-Attention} (TCA) to automatically learn the semantic correspondence of different semantic categories by introducing the proxy embedding.
	We further use such proxy embedding to interact with the feature representations via the 
	cross-attention mechanism.
	By introducing auxiliary loss for the updated proxy embedding, we find that it could effectively improve feature correlations of the same category and the feature discriminability between different classes.

	In summary, the main contributions of this paper are three folds. 
	(1) We propose the MCTrans, which constructs cross-scale contextual dependencies and appropriates semantic relationships for accurate biomedical segmentation. 
	(2) A novel learnable proxy embedding is introduced to build category dependencies and enhance feature representation through self-attention and cross-attention, respectively.  
	(3) We plug the designed MCTrans into a UNet-like network and evaluate its performance on the six challenging segmentation datasets.
	The experiments show that MCTrans outperforms state-of-the-art methods by a significant margin with a slight computation increase in all tasks.
	%
	%
	These results demonstrate the effectiveness of all proposed network components.

	\section{Related Work}
	\hspace{1em}
	\textbf{Attention Mechanisms.} 
	Attention mechanisms have recently been used to construct pixel-level contextual representations.
	In specific, Oktay \textit{et al.}~\cite{oktay2018attention}  introduce an attention-based gate function to focus on the target and suppress irrelevant background.
	Lei \textit{et al.}~\cite{mou2019cs} further incorporate the feature-channel attention to model contextual dependencies in a more comprehensive manner.
	Moreover, Wang \textit{et al.}~\cite{wang2018non} propose the non-local operations to connect each pair of pixels to accurately model their relationship.
	These methods establish context by modeling the semantic and spatial relationships between pixels in a single scale but neglect more rich information presented in other scales.
	In this paper, we utilize the transformer's power to construct pixel-level contextual dependencies between multiple-scale features, enabling flexible information exchange across different scales and producing more appropriate visual representations.
	
	\textbf{Transformer.}
	The Transformer was proposed by Vaswani \textit{et al.}~\cite{vaswani2017attention}  and first applied in the machine translation, which performs information exchange between all pairs of the inputs via the self-attention mechanism. 
	Recently, Transformer has been proven its power in many computer vision tasks, including image classification \cite{dosovitskiy2020image}, semantic segmentation~\cite{xie2021segmenting}, object detection and tracking~\cite{zhu2020deformable,carion2020end}, and so on.
	For medical image segmentation, our concurrent work TransUnet~\cite{chen2021transunet} employs Transformer-Encoder on the highest-level feature of UNet to collect long-range dependencies.
	Nonetheless, the methods mentioned above are not specifically designed for medical image segmentation.
	Our work focuses on carefully developing a better transformer-based approach, thoroughly leveraging the attention mechanism's advantages for medical image segmentation.
	
	\section{Multi-Compound Transformer Network}
	As illustrated in Fig. \ref{fig:overview}, we introduce the MCTransformer between the classical UNet encoder and decoder architectures, which consists of the Transformer-Self-Attention (TSA) module and Transformer-Cross-Attention (TCA) module.
	The former is introduced to encode the contextual information between the multiple features, yielding rich and consistent pixel-level context. 
	And the latter introduces learnable embedding for semantic relationship modeling and further enhances feature representations.

	In practice, given an image $I \in \mathbb{R}^{H \times W} $, a deep CNN is adopted to extract multi-level features with different scales $\left\{X_i \in  \mathbb{R}^{ \frac{H}{2^i} \times \frac{W}{2^i} \times C_i} \right\}$. 
	For level $i$, features are unfolded with patch size of $P \times P$, where $P$ is set to 1 in this paper,  that is, each location of the $i$-th feature map will be considered as the "patch", yielding total $L_i = \frac{HW}{2^{2*i} \times P^2}$ patches.
	Next, different level of split patches are passed through to individual projection heads (i.e. 1 $\times$ 1 convolution layer) with the same output feature dimension $C_e$ and attain the embedded tokens $T_i \in \mathbb{R}^{L_i \times C_e}$.
	In this paper, we concatenate the features of  $i=2,3,4$ level and form overall tokens $T\in \mathbb{R}^{L \times C}$, where $L=\sum_{i=2}^{4} L_i$.
	To compensate for missing position information, positional embedding $E_{pos} \in \mathbb{R}^{L \times C}$  is supplemented to the tokens to provide information about the relative or absolute position of the feature in the sequence, which can be formulated as $T=T+E_{pos}$.
	Next, we feed the tokens into the TSA module for multi-scale context modeling. The output enhanced tokens are further pass through the TCA module and interact with the proxy embedding $E_{pro}  \in \mathbb{R}^{M\times C}$, where $M$ is the number of categories of the dataset.
	Finally, we fold the encoded tokens back to pyramid features and merge them in a bottom-up style to obtain the final feature map for prediction.

	\begin{figure}[t!]
		\centering
		\includegraphics[width=\linewidth]{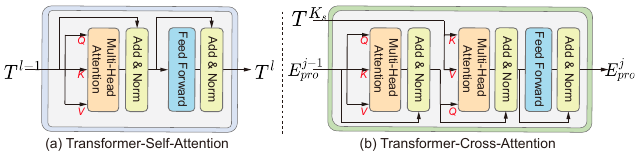}
		\caption{Illustration of the Transformer-Self-Attention  and Transformer-Cross-Attention modules.}
		\label{fig:attention}
	\end{figure}
	
	\subsection{Transformer-Self-Attention}
	Given the 1D embedding tokens $T$ as input, the TSA modules are employed to learn pixel-level contextual dependencies among multiple-scale features. 
	As illustrated in Fig. \ref{fig:overview}, the TSA module consists of  $K_s$ layers, each of which consists of multi-head self-attention (MSA) and feed forward networks (FFN) (see Fig. \ref{fig:attention} (a)), layer normalization (LN) is applied before every block and residual connection after every block. The FFN contains two linear layers with a ReLU activation. 
	For the $l$-th layers, the input to the self-attention is a triplet of (query, key, value) computed from the input $T^{l-1}$ as: 
	
	\begin{equation}
	\text {query}=T^{l-1} \mathbf{W}_{Q}^{l}, \text {key}=T^{l-1} \mathbf{W}_{K}^{l}, \text {value}=T^{l-1} \mathbf{W}_{V}^{l}
	\label{eq:1}
	\end{equation}
	where $\mathbf{W}_{Q}^{l}  \in \mathbb{R}^{C \times d_q}$, $\mathbf{W}_{K}^{l}  \in \mathbb{R}^{C \times d_k}$, $\mathbf{W}_{V}^{l}  \in \mathbb{R}^{C \times d_v}$  is the parameter matrices of different linear projections heads of $l$-th layer, and the $d_q$ , $d_k$, $d_v$ is the dimensions of three inputs. Self-Attention (SA) is then formulated as:
	
	\begin{equation}
	\operatorname{SA}\left(T^{l-1}\right)=T^{l-1}+\operatorname{Softmax}\left(\frac{T^{l-1} \mathbf{W}_{Q}^{l}\left(T^{l-1} \mathbf{W}_{K}^{l}\right)^{\top}}{\sqrt{d_k}}\right)\left(T^{l-1} \mathbf{W}_{V}^{l}\right)
	\label{eq:2}
	\end{equation}
	MSA is an extension with $h$ independent SA operations and project their concatenated outputs as: 
	\begin{equation}
	\operatorname{MSA}(T^{l-1})=\operatorname{Concat}\left(\operatorname{SA}_{1}, \ldots, \text { SA }_{\mathrm{h}}\right) W_{O}^{l}
	\label{eq:3}
	\end{equation}
	where $\mathbf{W}_{O} \in \mathbb{R}^{h d_k \times C}$ is a parameter of output linear projection head. In this paper, we employ $h=8$, $C=128$ and $d_{q}$,$d_{k}$,$d_{v}$ are equal to $C/h=32$.
	As depicted  in Fig. \ref{fig:attention} (a), the whole calculation can be formulated as:
	\begin{equation}
	T^{l}=\operatorname{MSA}\left(T^{l-1}\right)+\operatorname{FFN}\left(\operatorname{MSA}\left(T^{l-1}\right)\right) \in \mathbb{R}^{L \times C}
	\label{eq:4}
	\end{equation}
	We omitted the LN in the equation for simplicity.
	It should be noted the token $T$ (flatten from multi-scale features) has an extremely long sequence length, and the quadratic computation complexity of MSA  makes it not possible to handle.
	To this end, in this module, we use the Deformable Self Attention (DSA) mechanism proposed in \cite{zhu2020deformable} to replace the SA.
	As data-dependent sparse attention, which is not all-pairwise, DSA only attends to a sparse set of elements from the whole sequence regardless of its sequence length, which largely reduces computation complexity and allows the interactions of multi-level feature maps. For more details please refer to \cite{zhu2020deformable}. 

	\subsection{Transformer-Cross-Attention}
	
	As figured in Fig. \ref{fig:overview}, beside the enhanced tokens $T^{K_s}$, a learnable proxy embedding $E_{pro}$ is proposed to learn the global semantic relationship (i.e. intra-/inter- class) between categories.
	Like the TSA module, the TCA module consists of  $K_c$ layers but contains two multi-head self-attention blocks.
	In practice, for the $j$-th layer, the proxy embedding $E_{pro}^{j-1}$ is transformed by various linear projection heads to yield inputs (query, key, value) of the first MSA block.
	Here, the MSA block's self-attention mechanism connects and interacts with each pair of categories, thus modeling the semantic correspondence of various labels.
	Next, the learned proxy embedding extracts and interacts with the features of the input tokens $T^{K_s}$ via the cross attention in another MSA block, where the query input is the proxy embedding, key, and value inputs are the tokens $T^{K_s}$.
	Through the cross-attention, the features of tokens communicate with the learned global semantic relationship, comprehensively improving intra-class consistency and the inter-class discriminability of feature representation, yielding updated proxy embedding $E_{pro}^{j}$. 
	Noted that the calculation of procedure two MSA block is equal to Eq. \ref{eq:2}.
	Moreover, we introduce an auxiliary loss $Loss_{aux}$ to promote proxy embedding learning.
	In particular, the output $E_{pro}^{K_c}$of the last layer of the TCA module is further passed to a linear projection head and yields a multi-class prediction $Pred_{aux} \in \mathbb{R}^{M} $. Base on the ground-truth segmentation mask, we find the unique elements to compute classification labels for supervision.
	In this way, the proxy embedding is driven to learn appropriate semantic relationship, and help to improve feature correlations of the same category and the feature discriminability between different categories.
	Finally, the encoded tokens $T^{K_s}$  is fold back to 2D features and append the uninvolved features to form the pyramid features $\left\{X_{0}, X_{1}, X_{2}^{\prime}, X_{3}^{\prime}, X_{4}^{\prime} \right\}$. We merge them progressively in regular bottom-up style with a 2$\times$ upsampling layer and a $3\times3$ convolution to attain the final feature map for segmentation. For more details of the construction of multi-scale feature maps, please refer to Appendix.	
	\setlength{\tabcolsep}{2.75pt}
	\renewcommand{\arraystretch}{0.85}
	\begin{table}[t!]
		\small
		\begin{center}
			\begin{tabular}{l||cc||cccccc}
				\hline
				Method               &   Params (M)   &    GFlops & Neo & Inflam & Conn&Dead & Epi & Ave\\
				\hline\hline
				UNet \cite{ronneberger2015u}&7.853&14.037 & 82.86&66.16&62.45&38.10&75.02&64.92 \\
				UNet \cite{ronneberger2015u}+NonLocal \cite{wang2018non} & 8.379 & 14.172 &82.67&67.48&62.63&40.44&76.41&65.93 \\
				UNet \cite{ronneberger2015u}+VIT-Enc \cite{dosovitskiy2020image} & 27.008 &  18.936 & 83.34&68.33&63.18&38.11&77.25&66.04 \\
				MCTrans w/o TCA &7.115 &18.061&83.87&68.54&64.68&44.25&78.30&67.93\\
				MCTrans w/o TSA &6.167  & 11.589&83.39&67.82 &63.94 &44.35 &76.31 &67.16\\
				MCTrans w/o Aux-Loss &7.642&18.065&83.92&67.92 &64.22 &45.16 &78.14 &67.87\\
				\rowcolor{gray!15} 
				MCTrans  &7.642&18.065&83.99&68.24&64.95&46.39&78.42&68.40\\
				\hline
			\end{tabular}
		\end{center}
		\caption{Ablation studies of core components of MCTrans. The performance is evaluated on Pannuke dataset. We estimate Flops and parameters by using [1$\times$3$\times$256$\times$256] input. Note that, UNet+VIT-Enc network is equivalent to TransUNet.}
		\vspace{-0.3in}
		\label{tab:ablation_study}
	\end{table}
	\vspace{-0.15in}
	\section{Experiments}
	\subsection{Datasets and Settings}
	The proposed MCTrans was evaluated on six segmentation datasets of three types. 
	(1) Cell Segmentation \cite{gamper2019pannuke}:
	Pannuke dataset (pathology, 7,904 cases, 6 classes),
	(2) PolyP Segmentation \cite{bernal2015wm,bernal2012towards,silva2014toward,jha2020sessile}: CVC-Clinic dataset (colonoscopy, 612 cases, 2 classes), CVC-ColonDB dataset (380 cases, 2 classes), ETIS-Larib dataset (196 cases, 2 classes), Kvasir dataset (1,000 cases, 2 classes), 
	(3) Skin Lesion Segmentation \cite{codella2019skin}: ISIC2018 dataset (dermoscopy, 2,594 cases, 2 classes).
	Each task has different data modalities, data sizes, and foreground classes, making them suitable for evaluating the effectiveness and generalization of the MCTrans.
	For cell segmentation, we report the results of the officially divided 3-fold cross-validation.
	For other tasks, since the annotation of test set is not publicly available, we report the 5-fold cross-validation results.
	Below, we mainly evaluate our approach on the Panunke dataset to show the effectiveness of different network components. Finally, we compare our MCTrans with the top methods on all of the datasets.
	We report all results in terms of the Dice Similarity Coefficient  (DSC), and a better score indicates a better result.
	
	We construct the MCTrans with the PyTorch toolkit. We adopt conventional CNN backbone networks, including VGG-Style \cite{simonyan2014very}  encoder and ResNet-34 \cite{he2016deep}, to extract multi-scale feature representations.
	For network optimization, we use the cross-entropy loss and dice loss to penalize the training error of segmentation and a cross-entropy loss with a weight of 0.1 for auxiliary supervision. 
	We augment the training images with simple flipping. We use the Adam optimizer with an initial learning rate of 3e-4 to train the network. The learning rate is decayed linearly during the training. All models are trained on 1 V100 GPU. Please refer to the Appendix for more training details of specific datasets.

	\begin{figure}[t!]
		\centering
		\includegraphics[width=0.98\linewidth]{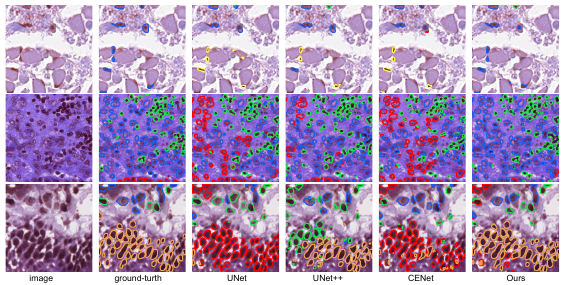}
		\vspace{-0.1in}
		\caption{Segmentation results on the Pannuke dataset, which contains of five foreground classes: \textcolor{Neoplastic}{Neoplastic}, \textcolor{Inflammatory}{Inflammatory},   \textcolor{Connective}{Connective},
			\textcolor{Dead}{Dead}, and \textcolor{Non-Neoplastic Epithelial}{Non-Neoplastic Epithelial}.}
		\label{fig:visual_results}
		\vspace{-0.15in}
	\end{figure}
	\subsection{Ablation Studies}
	\subsubsection{Analysis of the Network Components}
	We evaluate the importance of the core modules of MCTrans by the segmentation accuracy.
	We use the VGG-style network as the backbone.
	Compared to the UNet baseline which achieves a 64.92\% dice score on the Pannuke dataset, MCTrans use TSA and TCA's power to achieve the accuracy of 68.40\%.
	In Table. \ref{tab:ablation_study}, the performance is promoting to 67.93\% by adding the TSA module to the Unet. 
	To demonstrate the effectiveness of constructing multiple-scale pixel-level dependency, we employ the Non-local operation and Transformer-Encoder \cite{dosovitskiy2020image} on UNet's highest levels features to enable single-scale context propagation, yielding accuracies far behind our method.
	We further evaluate the influence of the TCA module. After adding the TCA, the learned semantic prior help to construct identified context dependencies and improve the score of Baseline and MCTrans to 67.16\% and 68.40\%, respectively. 
	It indicates the effectiveness of learning semantic relationships to enhance the feature representations.
	We also investigate the case of removing auxiliary loss. Here, we only model semantic relationships among categories implicitly. This strategy degrades the performance to 67.87\%.
	\vspace{-0.2in}
	\setlength{\tabcolsep}{7.25pt}
	\renewcommand{\arraystretch}{0.8}
	\begin{table}[h]
		\small
		\begin{center}
			\begin{tabular}{ccccc||ccccc} \hline
				$N_s$ & 2 & 4 & 6 & 8 & $N_c$ & 2 & 4 & 6 & 8 \\ 
				\hline \hline
				DSC  &67.25  & 67.67&\textbf{67.93}   & 67.50   & -  & 68.15 &\textbf{68.40}   &68.31   & 68.11  \\ \hline
			\end{tabular}
		\end{center}
		\vspace{-0.05in}
		\caption{Sensitivity to the number of the TSA and TCA module.}
		\label{tab:sense_number}
		\vspace{-0.75in}
	\end{table}
	
	\subsubsection{Sensitivity to the Setting}
	We change the number of TSA and TCA modules and study the effect on the segmentation accuracy.
	We first increase the number $N_s$ of the TSA module gradually to enlarge the modeling capacity. 
	As shown in Table. \ref{tab:sense_number}, we can see that when the size of TSA increases, the DSC score first increases and then decreases.
	After fixing  $N_s$, we further plug the TCA and enlarge its size.
	We also discover that it reaches the top at $N_c=4$ and then decreases.
	This indirectly shows that the capacity of transformer-based model is not as large as better when training on a small dataset.

	\subsection{Comparisons with State-of-the-art Methods}
	
	In Table. \ref{tab:comparsion_with_sota_0}, we compare the MCTrans with the state-of-the-art methods on the Pannuke dataset.
	In the first group,  we adopt a conventional VGG-Style network as feature extractor. Compared to other modeling mechanisms, our MCTrans achieves significant improvement by investing pixel-level dependencies across multiple-levels features.
	For a more comprehensive comparison, in the second group, we adopt a stronger features extractor (e.g., ResNet-34). Again, we achieve better accuracies than other methods. 
	We provide the examples of the segmentation results in Fig. \ref{fig:visual_results}. In Table. \ref{tab:comparsion_with_sota_1}. We also report the results on five lesion segmentation, respectively. The results of our method still outperform other top methods by a significant margin.
	Such results demonstrate the versatility of the proposed MCTrans on various segmentation tasks.

	We provide more details of the computational overheads (i.e. floating-point operations per second (Flops) and the number of parameters).
	As shown in Table.~\ref{tab:comparsion_with_sota_0}, MCTrans achieves better results at the cost of reasonable computational overheads.
	Compared to the UNet baseline, MCTrans with almost identical parameters and a slight computation increase achieves a significant improvement of 3.64\%.
	Note that the other top methods, such as UNet++, surpass MCTrans over much computation while yielding lower performance.
	
	\setlength{\tabcolsep}{3.5pt}
	\renewcommand{\arraystretch}{0.85}
	\begin{table}[t!]
		\small
		\begin{center}
			\begin{tabular}{l||cc||cccccc}
				\hline
				Method &   Params (M)   &    Flops (G) & Neo & Inflam & Conn&Dead & Epi & Ave\\
				\hline\hline
				UNet \cite{ronneberger2015u}&7.853&14.037 & 82.86&66.16&62.45&38.10&75.02&64.92 \\
				UNet++ \cite{zhou2018unet++}&9.163&34.661 &82.14 & 66.01 & 61.61 &38.47 &76.54 & 64.97\\
				CENet \cite{gu2019net}&17.682&18.779 &83.05& 66.92& 62.41&38.021&76.44&65.37\\
				AttentionUNet \cite{oktay2018attention}&8.382&15.711  &81.85& 65.37& 63.79&38.96&75.45&64.27 \\
				\rowcolor{gray!15} 
				MCTrans  &7.642&18.065&83.99&68.24&63.95&47.39&78.42&68.40\\
				\hline\hline
				UNet \cite{ronneberger2015u}&24.563&38.257 &82.85& 65.48  &62.29 &40.11&75.57& 65.26   \\
				UNet++ \cite{zhou2018unet++}&25.094&84.299&82.03& 67.58  & 62.79&40.79 &77.21& 66.08   \\
				CENet \cite{gu2019net}&34.368&41.389 &82.73& \textbf{68.25} & 63.15 &41.12 &77.27&66.50 \\
				AttentionUNet \cite{oktay2018attention}&25.094&40.065 &82.74& 65.42  & 62.09 &38.60 &76.02 & 64.97  \\
				\rowcolor{gray!15} 
				MCTrans &23.787&39.71 &\textbf{84.22} &68.21 &\textbf{65.04} &\textbf{48.30} &\textbf{78.70} &\textbf{68.90}\\
				\hline
			\end{tabular}
		\end{center}
		\vspace{-0.1in}
		\caption{Comparisons with other conventional methods on the Pannuke dataset.}
		\vspace{-0.2in}
		\label{tab:comparsion_with_sota_0}
	\end{table}
	\setlength{\tabcolsep}{8.2pt}
	\renewcommand{\arraystretch}{0.85}
	\begin{table}[t!]
		\small
		\begin{center}
			\begin{tabular}{l||ccccc}
				\hline
				Method & CVC-Clinic & CVC-Colon & ETIS &Kavairs & ISIC2018 \\
				\hline\hline
				UNet \cite{ronneberger2015u}&88.59 &82.24 &80.89  &84.32&88.78\\
				UNet++ \cite{zhou2018unet++} &89.30  &82.86&80.77   &84.95&88.85 \\
				CENet \cite{gu2019net}  &91.53  &83.11 & 75.03  &84.92 &89.53\\
				AttentionUNet \cite{oktay2018attention}& 90.57  &83.25 &79.68 & 80.25 &88.95\\
				\rowcolor{gray!15} 
				MCTrans&\textbf{92.30}&\textbf{86.58}&\textbf{83.69}&\textbf{86.20}&\textbf{90.35}\\
				\hline
			\end{tabular}
		\end{center}
		\caption{Comparisons with other top methods on the five lesion segmentation datasets.}
		\vspace{-0.4in}
		\label{tab:comparsion_with_sota_1}
	\end{table}
	
	\section{Conclusions}
	In this paper, we propose a powerful transformer-based network for medical image segmentation.
	Our method incorporates rich context modeling and semantic relationship mining via powerful attention mechanisms, effectively address the issues of cross-scale dependencies, the semantic correspondence of different categories, and so on. 
	Our approach is effective and outperforms the state-of-the-art method such as TransUnet on several public datasets.
	
	\section*{Acknowledgments}	
	This work is partially supported by the General Research Fund of Hong Kong No. 27208720, the Open Research Fund from Shenzhen Research Institute of Big Data No. 2019ORF01005, and the Research Donation from SenseTime Group Limited, the NSFC-Youth 61902335 and SRIBD Open Funding,  the founding of Science and Technology Commission Shanghai Municipality  No.19511121400.

	\bibliographystyle{splncs04}
	\bibliography{library}

\begin{thebibliography}{10}
\providecommand{\url}[1]{\texttt{#1}}
\providecommand{\urlprefix}{URL }
\providecommand{\doi}[1]{https://doi.org/#1}

\bibitem{bernal2015wm}
Bernal, J., S{\'a}nchez, F.J., Fern{\'a}ndez-Esparrach, G., Gil, D.,
  Rodr{\'\i}guez, C., Vilari{\~n}o, F.: Wm-dova maps for accurate polyp
  highlighting in colonoscopy: Validation vs. saliency maps from physicians.
  Computerized Medical Imaging and Graphics  \textbf{43},  99--111 (2015)

\bibitem{bernal2012towards}
Bernal, J., S{\'a}nchez, J., Vilarino, F.: Towards automatic polyp detection
  with a polyp appearance model. Pattern Recognition  \textbf{45}(9),
  3166--3182 (2012)

\bibitem{carion2020end}
Carion, N., Massa, F., Synnaeve, G., Usunier, N., Kirillov, A., Zagoruyko, S.:
  End-to-end object detection with transformers. In: European Conference on
  Computer Vision. pp. 213--229. Springer (2020)

\bibitem{chen2021transunet}
Chen, J., Lu, Y., Yu, Q., Luo, X., Adeli, E., Wang, Y., Lu, L., Yuille, A.L.,
  Zhou, Y.: Transunet: Transformers make strong encoders for medical image
  segmentation. arXiv preprint arXiv:2102.04306  (2021)

\bibitem{chen2017deeplab}
Chen, L.C., Papandreou, G., Kokkinos, I., Murphy, K., Yuille, A.L.: Deeplab:
  Semantic image segmentation with deep convolutional nets, atrous convolution,
  and fully connected crfs. IEEE transactions on pattern analysis and machine
  intelligence  \textbf{40}(4),  834--848 (2017)

\bibitem{codella2019skin}
Codella, N., Rotemberg, V., Tschandl, P., Celebi, M.E., Dusza, S., Gutman, D.,
  Helba, B., Kalloo, A., Liopyris, K., Marchetti, M., et~al.: Skin lesion
  analysis toward melanoma detection 2018: A challenge hosted by the
  international skin imaging collaboration (isic). arXiv preprint
  arXiv:1902.03368  (2019)

\bibitem{dosovitskiy2020image}
Dosovitskiy, A., Beyer, L., Kolesnikov, A., Weissenborn, D., Zhai, X.,
  Unterthiner, T., Dehghani, M., Minderer, M., Heigold, G., Gelly, S., et~al.:
  An image is worth 16x16 words: Transformers for image recognition at scale.
  arXiv preprint arXiv:2010.11929  (2020)

\bibitem{gamper2019pannuke}
Gamper, J., Koohbanani, N.A., Benet, K., Khuram, A., Rajpoot, N.: Pannuke: an
  open pan-cancer histology dataset for nuclei instance segmentation and
  classification. In: European Congress on Digital Pathology. pp. 11--19.
  Springer (2019)

\bibitem{gu2019net}
Gu, Z., Cheng, J., Fu, H., Zhou, K., Hao, H., Zhao, Y., Zhang, T., Gao, S.,
  Liu, J.: Ce-net: Context encoder network for 2d medical image segmentation.
  IEEE transactions on medical imaging  \textbf{38}(10),  2281--2292 (2019)

\bibitem{he2016deep}
He, K., Zhang, X., Ren, S., Sun, J.: Deep residual learning for image
  recognition. In: Proceedings of the IEEE conference on computer vision and
  pattern recognition. pp. 770--778 (2016)

\bibitem{jha2020sessile}
Jha, D., Smedsrud, P.H., Johansen, D., de~Lange, T., Johansen, H.D., Halvorsen,
  P., Riegler, M.A.: A comprehensive study on colorectal polyp segmentation
  with resunet++, conditional random field and test-time augmentation (2020)

\bibitem{ji2020uxnet}
Ji, Y., Zhang, R., Li, Z., Ren, J., Zhang, S., Luo, P.: Uxnet: Searching
  multi-level feature aggregation for 3d medical image segmentation. In:
  International Conference on Medical Image Computing and Computer-Assisted
  Intervention. pp. 346--356. Springer (2020)

\bibitem{long2015fully}
Long, J., Shelhamer, E., Darrell, T.: Fully convolutional networks for semantic
  segmentation. In: Proceedings of the IEEE conference on computer vision and
  pattern recognition. pp. 3431--3440 (2015)

\bibitem{mou2019cs}
Mou, L., Zhao, Y., Chen, L., Cheng, J., Gu, Z., Hao, H., Qi, H., Zheng, Y.,
  Frangi, A., Liu, J.: Cs-net: channel and spatial attention network for
  curvilinear structure segmentation. In: International Conference on Medical
  Image Computing and Computer-Assisted Intervention. pp. 721--730. Springer
  (2019)

\bibitem{oktay2018attention}
Oktay, O., Schlemper, J., Folgoc, L.L., Lee, M., Heinrich, M., Misawa, K.,
  Mori, K., McDonagh, S., Hammerla, N.Y., Kainz, B., et~al.: Attention u-net:
  Learning where to look for the pancreas. arXiv preprint arXiv:1804.03999
  (2018)

\bibitem{ronneberger2015u}
Ronneberger, O., Fischer, P., Brox, T.: U-net: Convolutional networks for
  biomedical image segmentation. In: International Conference on Medical image
  computing and computer-assisted intervention. pp. 234--241. Springer (2015)

\bibitem{silva2014toward}
Silva, J., Histace, A., Romain, O., Dray, X., Granado, B.: Toward embedded
  detection of polyps in wce images for early diagnosis of colorectal cancer.
  International journal of computer assisted radiology and surgery
  \textbf{9}(2),  283--293 (2014)

\bibitem{simonyan2014very}
Simonyan, K., Zisserman, A.: Very deep convolutional networks for large-scale
  image recognition. arXiv preprint arXiv:1409.1556  (2014)

\bibitem{vaswani2017attention}
Vaswani, A., Shazeer, N., Parmar, N., Uszkoreit, J., Jones, L., Gomez, A.N.,
  Kaiser, L., Polosukhin, I.: Attention is all you need. arXiv preprint
  arXiv:1706.03762  (2017)

\bibitem{wang2018non}
Wang, X., Girshick, R., Gupta, A., He, K.: Non-local neural networks. In:
  Proceedings of the IEEE conference on computer vision and pattern
  recognition. pp. 7794--7803 (2018)

\bibitem{xie2021segmenting}
Xie, E., Wang, W., Wang, W., Sun, P., Xu, H., Liang, D., Luo, P.: Segmenting
  transparent object in the wild with transformer. arXiv preprint
  arXiv:2101.08461  (2021)

\bibitem{yu2020context}
Yu, C., Wang, J., Gao, C., Yu, G., Shen, C., Sang, N.: Context prior for scene
  segmentation. In: Proceedings of the IEEE/CVF Conference on Computer Vision
  and Pattern Recognition. pp. 12416--12425 (2020)

\bibitem{zhao2017pyramid}
Zhao, H., Shi, J., Qi, X., Wang, X., Jia, J.: Pyramid scene parsing network.
  In: Proceedings of the IEEE conference on computer vision and pattern
  recognition. pp. 2881--2890 (2017)

\bibitem{zhou2018unet++}
Zhou, Z., Siddiquee, M.M.R., Tajbakhsh, N., Liang, J.: Unet++: A nested u-net
  architecture for medical image segmentation. In: Deep learning in medical
  image analysis and multimodal learning for clinical decision support, pp.
  3--11. Springer (2018)

\bibitem{zhu2020deformable}
Zhu, X., Su, W., Lu, L., Li, B., Wang, X., Dai, J.: Deformable detr: Deformable
  transformers for end-to-end object detection. arXiv preprint arXiv:2010.04159
   (2020)

\end{thebibliography}

\end{document}